\documentclass[10pt,twocolumn,letterpaper]{article}

\usepackage{cvpr}              % To produce the CAMERA-READY version
\usepackage{booktabs}
\usepackage{array}
\usepackage{booktabs}
\usepackage{tabularx}
\usepackage{xcolor}
\usepackage{soul}      % for \hl highlighting
\usepackage{fancyvrb}  % optional, only if you want verbatim blocks
\usepackage{graphicx}
\usepackage[export]{adjustbox}
\usepackage[accsupp]{axessibility}

\definecolor{hlblue}{RGB}{209,226,246}
\sethlcolor{hlblue}

\newcolumntype{Y}{>{\raggedright\arraybackslash}X}

\definecolor{cvprblue}{rgb}{0.21,0.49,0.74}
\usepackage[pagebackref,breaklinks,colorlinks,allcolors=cvprblue]{hyperref}
\usepackage{booktabs,tabularx}

\usepackage[normalem]{ulem} % Provides strikeout (sout) command

\def\paperID{*****} % *** Enter the Paper ID here
\def\confName{CVPR}
\def\confYear{2026}

\title{A Human-in-the-Loop Framework for Efficient Prompt Selection in Microscopy Vision–Language Models}

\author{
Abhiram Kandiyana$^{1}$\quad
Ankur Mali$^{1}$ \quad
Lawrence O. Hall$^{1}$ \quad
Peter R. Mouton$^{2}$ \quad 
Dmitry Goldgof$^{1}$\\
{\small $^{1}$University of South Florida, USA \quad
$^{2}$SRC Biosciences, USA} \\
{\tt\small \{kandiyana, ankurarjunmali, lohall, goldgof\}@usf.edu \quad peter@disector.com}
}

\begin{document}
\maketitle
\begin{abstract}
 Deep-learning pipelines for microscopy image classification often require expensive, labor- and time-intensive expert annotation to produce high-quality ground truth for training. Recent work has shown that prompt tuning of vision--language models (VLMs) can reduce manual annotation by constructing a small prompt set of expert-verified image--caption exemplars that is reused as few-shot context to classify all remaining images at inference time. To further reduce effort, the VLM can draft captions for candidate exemplars, which experts then verify and lightly edit instead of writing text de novo. However, two practical questions remain unaddressed: (1) which unlabeled images should be prioritized for verification, and (2) how many verified exemplars are needed to reach a performance target.

In this work, we address these questions by formulating prompt-set construction as a target-driven active learning problem that prioritizes which images to annotate.

We study three complementary selection criteria under strict low-resource constraints with small unlabeled pools. Experiments show that our methods reach the target performance with substantially fewer expert-verified images than random selection, achieving $100\%$ test accuracy with as few as \(\sim\)20 annotated images on average.

More broadly, Our human-in-the-loop framework demonstrates a human-centered use of generative AI in biomedical image analysis, where experts remain actively involved in verifying and refining model output while significantly reducing annotation cost. \emph{Code and data will be publicly available}\footnote{https://abhiram-kandiyana.github.io/APT/}.
% \footnote{Corresponding author: lohall@usf.edu}
\end{abstract}    
\section{Introduction}

Accurate assessment of cellular damage is essential for studying brain aging, neuroinflammation, neurotoxicology, and the efficacy of treatments for neurological disorders such as Parkinson’s, Huntington’s, and Alzheimer’s diseases. In many microscopy-based studies, though multiple images are collected per animal (case), labels are ultimately assigned at the \emph{animal level} because the clinically and biologically meaningful outcome is a single diagnosis per animal (e.g., control or treated), making the task naturally aligned with a \emph{multiple-instance} learning setting \cite{wsi-mil}. 

Traditional quantification using manual counting under unbiased stereology rules is a highly accurate and precise method for cellular damage analysis, but it is extremely labor- and time-intensive for domain experts \cite{stereo-book, mouton2014neurostereology}.

Prior work has shown that deep learning models can automate parts of this workflow, including quantifying cell loss within defined regions of interest using disector-based multiple-input multiple-output (MIMO) frameworks and related CNN-based approaches on low-magnification images \cite{palakmimounet,morera2023mimo,morera20x}. While these methods can achieve strong performance and improved throughput relative to manual counting, they often require substantial amounts of expert-generated ground truth, e.g., manual counts, extensive image capture to overcome biological variation, and non-trivial model development effort.

A key limitation of these methods is that high-quality annotation in biomedical microscopy is expensive and difficult to scale. It requires trained personnel and demands significant expert supervision. Moreover, training and fine-tuning deep learning models can take weeks to months and must often be repeated when datasets, staining protocols, imaging conditions, or target phenotypes change. 
% These constraints motivate methods that can achieve a predefined target performance with minimal expert annotation.

Large Vision--Language Models (VLMs) offer a promising alternative by enabling \emph{in-context learning}: a frozen pre-trained model can be adapted at inference time by prepending a small set of exemplars (prompt set), without gradient-based fine-tuning \cite{why-ICL}. In addition to producing class predictions, many VLMs can generate natural-language rationales that describe salient visual evidence supporting the decision, providing a form of interpretable, domain-aligned explanation that is not naturally available in standard image classification pipelines.

However, prompt tuning introduces its own bottleneck: performance depends strongly on the quality and composition of the prompt set included in the input context \cite{ICL-examples}. In biomedical settings, constructing a prompt set is itself costly because it requires expert-written labels and captions. A fixed exemplar set size also leaves open two practical questions: 1) Which unlabeled images are most valuable to annotate for a given task? 2) How many annotated images are actually necessary to achieve a task-specific performance target?

In this work, we frame prompt set construction as a target-driven Active Learning (AL) problem for VLM-based microscopy classification. Starting from a very small seed prompt set (10 images), we iteratively select a small active set of unlabeled images, use the VLM to draft captions under the current prompt context, and have an expert verify and correct these captions. The corrected exemplars are appended to the prompt set, and the loop terminates as soon as a validation performance target is achieved. Within this framework, the central design choice is the \emph{acquisition function} that selects which images are most valuable to annotate next.

We study three acquisition functions that differ in how the active set is selected: (i) uncertainty-guided acquisition using stochastic decoding, (ii) complexity-aware uncertainty acquisition that favors informative yet concise and non-redundant exemplars, and (iii) density-tree boundary sampling that targets informative regions in the image encoder embedding space. On a biomedical microscopy dataset, the proposed acquisition functions reach the validation target with only 20 expert-verified annotations and achieve 100\% animal-level test accuracy, demonstrating strong annotation efficiency under realistic low-resource conditions.
% \JMBcomment{Few dozen implies greater than 36 annotations.}

Our main contributions are:
\begin{enumerate}\setlength\itemsep{2pt}
    \item A target-driven and annotation-efficient AL-based prompt tuning framework for microscopy image classification.
    \item Three complementary acquisition functions that capture distinct selection principles (model uncertainty, exemplar complexity, and embedding-space structure), enabling robust annotation-efficient prompting across different low-resource constraints.
    \item Empirical evidence that acquisition-aware prompt tuning substantially improves annotation efficiency while maintaining strong test performance under low-resource constraints.
    % \item An ablation study showing how the embedding backbone influences density-tree boundary acquisition behavior.
\end{enumerate}
\section{Related Work}
\label{sec:related_work}

\subsection{Active Learning}
Our work builds on Active Learning (AL), which aims to reduce labeling cost by iteratively selecting the most informative unlabeled samples for annotation \cite{settles2009,lewis1994,tong2001}. In general, AL strategies are commonly grouped into \emph{uncertainty-based} methods \cite{entropy, margin_AL} which prioritize samples the model is least confident about (e.g., entropy, margin), \emph{diversity-based} methods \cite{coreset} which encourage coverage of diverse regions of the feature space, and hybrids that combine both criteria to avoid redundancy \cite{settles2009, BADGE}. 

While uncertainty and diversity are often effective in classical settings, recent analyses show that, on complex multimodal tasks, standard AL heuristics can sometimes fail to outperform (and may even underperform) random sampling. In particular, Karamcheti \emph{et al.} identify \emph{collective outliers} which are groups of samples that AL methods tend to acquire early because they appear highly uncertain, yet models fail to learn from them and demonstrate that this phenomenon can degrade pool-based AL relative to random selection \cite{collectiveOutliers}. This observation motivates incorporating safeguards that prevent selection from being dominated by small, isolated, ``collective outlier" regions of the pool.

% Recent work has proposed AL variants that explicitly aim to mitigate outlier-driven acquisition by combining uncertainty with local neighborhood structure. Safaei and Patel introduce \emph{Calibrated Entropy-weighted Clustering (CEC)}, which calibrates predictive entropy, incorporates neighbor-aware uncertainty to reduce emphasis on isolated hard examples, and then applies uncertainty-weighted \emph{k}-means clustering over the unlabeled pool to encourage diversity within each queried batch \cite{cec}. However, because CEC relies on centroid-based \emph{k}-means with pool-level uncertainty weighting, its selection behavior is governed by a global clustering objective whose stability can depend on the composition and size of the available unlabeled pool.

Recent work has proposed AL variants that mitigate outlier-driven acquisition by coupling uncertainty with neighborhood structure. \cite{cec} introduces \emph{Calibrated Entropy-weighted Clustering (CEC)}, which calibrates predictive entropy, incorporates neighbor-aware uncertainty, and applies uncertainty-weighted \emph{k}-means clustering over the unlabeled pool to promote diverse selection. However, because CEC is driven by a centroid-based global clustering objective with pool-level uncertainty weighting, its behavior depends on the composition and size of the available unlabeled pool. 
% This motivated us to explore acquisition rules based on local neighborhood that remain effective even when only a small unlabeled pool is available per round.

% This motivated us to explore an acquisition rule based on the local neighborhood and density structure that do not require fitting global centroids and remains effective when only a small unlabeled pool is available.
% Recent work has begun tailoring AL to vision--language models (VLMs). Safaei and Patel proposed a hybrid approach that calibrates predictive entropy, incorporates neighbor-aware uncertainty, and then performs uncertainty-weighted \emph{k}-means clustering of the unlabeled pool to promote diversity during selection \cite{cec}. On the other hand, Bang \emph{et al.} introduce an AL framework for adapting prompt-based VLM classifiers that addresses the inherent class imbalance in CLIP \cite{clip} via a pseudo-class balancing module that uses model predictions of the entire unlabeled pool to improve class balance within the active set, and can be combined with standard AI methods (e.g., entropy, coreset, BADGE) \cite{bang2024active}.

\subsection{Prompt Tuning}
\label{ssec:related_work-prompt_tuning}

Prompt tuning has emerged as an efficient adaptation strategy for VLMs, enabling them to specialize to downstream tasks or domain-specific distributions without full model fine-tuning. Rather than updating all model parameters, prompt tuning modifies the model input in a structured manner so that pretrained representations can be steered toward a target task during inference.

Broadly, prompt tuning methods can be categorized into two main paradigms: (1) prompt tuning in the embedding space \cite{coop,cocoop,residualprompttuning,bang2024active} and (2) prompt tuning in the language space \cite{LMMsForPathSPIE, active-prompt_acl_2024, gpt4vForNeuroScience,gpt4oISBI2025}. Although these approaches differ in implementation, their underlying principle is similar, task-relevant context is injected into the model input to guide predictions without altering the model parameters.

In embedding-space prompt tuning, a set of learnable context vectors are added to the input which are optimized via backpropagation in training while the core model weights remain frozen. This approach allows the model to adapt to new domains through compact, trainable prompt parameters, often resulting in parameter-efficient fine-tuning \cite{coop}. In contrast, language-space prompt tuning uses human-designed textual descriptions, task instructions, domain-specific context,  and/or few-shot exemplar demonstrations prepended to the input. By providing richer semantic context in natural language, the model can leverage its pretrained linguistic and multimodal knowledge to adapt to new tasks without modification to any weights. 

A closely related work is \emph{Active-Prompt} \cite{active-prompt_acl_2024}, which combines in-context learning with chain-of-thought (CoT) exemplars and an uncertainty-based selection strategy. Given a pool of unlabeled inputs, Active-Prompt repeatedly samples multiple model outputs to estimate prediction uncertainty, selects the most uncertain instances for human annotation of CoT rationales, and then uses the resulting annotated exemplars as few-shot context at inference time.

Beyond LLMs, recent work has shown that prompt-based adaptation can also be effective for downstream image classification by prepending exemplars to the input context of a frozen VLM at inference time \cite{LMMsForPathSPIE, gpt4vForNeuroScience, gpt4oISBI2025}. In particular, APT-USF \cite{gpt4oISBI2025} introduces a human-in-the-loop framework for microscopy image classification in which a \emph{prompt set} is manually captioned and the VLM is then used to generate captions for additional images that are subsequently verified and refined by an expert. This iterative process constructs an increasingly effective prompt set with substantially reduced manual overhead.

While APT-USF demonstrated that expert-verified, VLM-generated captions can be used to construct an effective prompt set for microscopy image classification, it also leaves open two practical questions relevant to annotation efficiency. First, the prompt set expansion step does not explicitly optimize \emph{which} unlabeled images are selected for expert verification. The unlabeled pool is sampled randomly, which can lead to redundant or less informative additions. Second, the number of images in the prompt set is treated largely as a fixed design choice. As a result, it is unclear whether comparable performance could be achieved with fewer image annotations under performance-based stopping criteria.

% Following these works, we likewise incorporate active learning into prompt-based adaptation of VLMs, but focus specifically on \emph{acquisition functions} that improve annotation efficiency under strict low-resource constraints for biomedical image classification tasks that require domain-informed reasoning. 

\section{Background}
\label{sec: background}

\subsection{Problem Formulation}
In this paper, the focus is to train a ${K}$-class  classifier while keeping the annotation cost as low as possible. This is done in an active learning setup in which the model interacts with an oracle (human-in-the-loop) that supplies ground-truth labels. In each round (${r}$), the model is trained using a labeled set, denoted as $\mathcal{D}_\ell=\{(x_i,y_i)\}_{i=1}^{L}\subset\mathcal{X}\times\mathcal{Y}$, where $\mathcal{X}$ denote the image space and $\mathcal{Y}=\{1,\dots,K\}$ the set of class labels. After training on $\mathcal{D}_\ell$, the active learning strategy queries the unlabeled pool $\mathcal{D}_u=\{x_j\}_{j=1}^{U}\subset\mathcal{X}$ to select $m$ informative samples per round for annotation. The oracle labels these queried instances, which are then added to $\mathcal{D}_\ell$ and used in subsequent training rounds. This training loop continues in rounds until a stopping criteria is met (annotation budget, target accuracy, max rounds etc).

\subsection{Prompt Tuning in Vision--Language Models}

 Prompt tuning in the language space as described in \cref{ssec:related_work-prompt_tuning} is used here. This choice is motivated by several considerations. First, language-based prompts offer greater interpretability, as the adaptation mechanism is expressed explicitly in natural language rather than in latent embedding parameters. Second, domain experts can directly encode task-specific knowledge, which is particularly valuable in biomedical imaging where input images often contain subtle, fine-grained morphological features that can be described precisely using expert terminology. By incorporating such rich descriptions and structured reasoning cues into the prompt, language-space prompting can more directly guide the model toward clinically or biologically meaningful visual evidence. Third, language-space prompting avoids the need for additional prompt parameter optimization, making it more practical in settings with limited labeled data and computational constraints.

We also incorporate explicit intermediate reasoning in the form of \emph{chain-of-thought} prompting, which encourages the model to produce step-by-step rationale before outputting a final decision \cite{COT}. This strategy has been shown to improve performance in large generative models, particularly on tasks that benefit from structured reasoning and compositional decision-making. In the context of VLMs, chain-of-thought style prompts are used to elicit explanations (e.g., identifying salient visual features, comparing candidate classes) prior to producing the final prediction. 

% We note that the specific formulation of chain-of-thought prompting used in this work is described in detail in a later section.

\section{Method}
\label{sec:method}

% \JMBcomment{Shouldn't the instances of $D_\ell$ and $\mathcal{D}_u$ below be calligraphic (to be consistent with Sec. 3.1)? Otherwise, $D_\ell$ and $\mathcal{D}_u$ are undefined.}
Given a small labeled set $\mathcal{D}_\ell$ and an unlabeled pool $\mathcal{D}_u$, our goal is to iteratively construct a \emph{prompt set} that serves as the evolving input context for the Vision--Language Model (VLM). Let $\mathcal{X}$ denote the image space and $\mathcal{Y}=\{1,\dots,K\}$ the set of class labels. We denote by $\mathcal{T}$ the space of captions, where a \emph{caption} $c \in \mathcal{T}$ is a free-form natural language description that (i) states the class label and (ii) provides a structured rationale for the decision, including chain-of-thought style reasoning that explains how the visual evidence supports the classification.

A \emph{prompt exemplar} is an image--caption pair $(x,c)\in \mathcal{X}\times\mathcal{T}$. At round $r$, the prompt set is defined as a collection of such exemplars,
\begin{equation}
\label{eq:prompt-set}
\mathcal{P}^{(r)}=\{(x_t^{(r)},c_t^{(r)})\}_{t=1}^{M_r}\subset \mathcal{X}\times\mathcal{T},
\end{equation}
% \JMBcomment{For consistency with how sets are defined, I would make $\mathcal{P}^{(r)}$ in equation (1) above also be calligraphic.}
which is prepended to the model input and used as few-shot context for inference and iterative refinement across rounds.

In addition to the prompt set, a fixed system prompt $S\subset\mathcal{T}$ provides task-level instructions, dataset context (e.g., imaging conditions), and output formatting constraints. Together, the system prompt and the current prompt set define the input context given to the model for generating predictions on validation images.

To achieve this, we adopt an iterative refinement process in rounds. At round $r$, the model receives images from an \emph{active set} $\mathcal{A}^{(r)} \subset \mathcal{D}_u$ together with the system prompt $S$ and the previous prompt set $\mathcal{P}^{(r-1)}$, and produces a predicted class along with an explanatory text for each image. An oracle (domain expert) verifies the predicted label and edits the explanation to ensure correctness, clarity, and alignment with domain knowledge. The corrected image--caption pairs are then appended to form the updated prompt set $\mathcal{P}^{(r)}$, and the corresponding images are removed from $\mathcal{D}_u$. Through this iterative augmentation, the prompt set progressively accumulates expert-validated exemplars that refine the model’s task adaptation.

A central component of this framework is the \emph{acquisition function}, which determines how the active set $\mathcal{A}^{(r)}$ of size $m$ is selected from $\mathcal{D}_u$ at each round. The acquisition function identifies the most informative images according to a principled criterion.

% This step is critical: given a fixed VLM, the choice of acquisition strategy directly governs how quickly the prompt set improves, how many rounds are required, and ultimately how many images must be annotated. 

In this section, we describe in detail the acquisition functions used to select the $m$ samples at each round.

\subsection{Uncertainty-Guided Acquisition}
\label{ssec:apt-u}

Previous studies have shown that model uncertainty can serve as an effective signal for selecting informative samples in Active Learning (AL) \cite{active-prompt_acl_2024, entropy}. 
% In both vision and language modeling settings, prioritizing uncertain instances has been demonstrated to accelerate performance gains while reducing annotation cost. More recently, uncertainty estimates derived from large language models have also been explored as a principled criterion for data selection. 
Motivated by these findings, we incorporate an uncertainty-based acquisition strategy within an iterative prompt-tuning framework for image classification.

At round $r$, for each image $x \in \mathcal{D}_u$, the frozen VLM receives the image together with the fixed system prompt $S$ and the prompt set from the previous round $\mathcal{P}^{(r-1)}$, and produces a predicted class $\hat{y}$ along with an explanatory caption $\hat{c}$. We estimate predictive uncertainty in a logit-free manner using stochastic decoding as shown in \cite{self-consistency}. Specifically, we generate $T$ independent stochastic outputs $\{(\hat{y}^{(t)}, \hat{c}^{(t)})\}_{t=1}^{T}$ by sampling from the model (via temperature parameter). From these outputs, we construct an empirical class distribution
\begin{equation}
p_y(x;\mathcal{P}^{(r-1)}) = \frac{1}{T} \sum_{t=1}^{T} \mathbf{1}\!\left[\hat{y}^{(t)} = y\right], \quad y \in \mathcal{Y},
\label{eq:eq-1}
\end{equation}
where $\mathbf{1}[\hat{y}^{(t)} = y]$ equals $1$ if the $t$-th stochastic prediction assigns class $y$, and $0$ otherwise.

We then quantify uncertainty using entropy,

{\small
\begin{equation}
u_{\text{ent}}(x,\mathcal{P}^{(r-1)}) = - \sum_{y=1}^{K} p_y(x,\mathcal{P}^{(r-1)}) \log p_y(x,\mathcal{P}^{(r-1)}),   
\label{eq: eq-2}
\end{equation}
}
where higher entropy indicates greater predictive uncertainty.

The active set $\mathcal{A}^{(r)}$ of size $m$ is formed by selecting the $m$ images in $\mathcal{D}_u$ with the highest uncertainty scores (${\rm Top-}m$), denoted as:
\begin{equation}
  \mathcal{A}^{(r)} = {\rm Top-}m \{ (x, u_{\text{ent}}(x;\mathcal{P}^{(r-1)})) : x \in \mathcal{D}_u \}.
\label{eq: eq-3}  
\end{equation}

\subsection{Complexity-Aware Uncertainty Acquisition}
\label{ssec:apt-ca}

Beyond uncertainty alone, we further account for the anticipated complexity of the captions that would be added to the prompt set. The motivation is that an informative exemplar should not only be uncertain under the current model state, but should also contribute concise and non-redundant explanatory content. To this end, we define a selection score that balances predictive uncertainty with expected caption complexity.

At round $r$, let $\mathcal{P}^{(r-1)}$ denote the current prompt set.
For an unlabeled image $x \in \mathcal{D}_u$, the frozen VLM receives 
$(x; S, \mathcal{P}^{(r-1)})$ and produces a predicted caption $\hat{c}(x;\mathcal{P}^{(r-1)}) \in \mathcal{T}$ together with a  label 
$\hat{y}(x;\mathcal{P}^{(r-1)})$.

Our goal is to select samples that are both
(i) uncertain under the current prompt set and
(ii) expected to contribute concise and non-redundant explanatory content.

Formally, this is posed as a constrained optimization problem:
\begin{equation}
\max_{x \in \mathcal{D}_u} 
\; u(x;\mathcal{P}^{(r-1)})
\quad 
\text{s.t.}
\quad 
\hat{\kappa}(x;\mathcal{P}^{(r-1)}) \le \tau,
\label{eq:constrained}
\end{equation}
where $u(\cdot)$ measures predictive uncertainty and 
$\hat{\kappa}(\cdot)$ measures anticipated caption complexity.
% \Abhiramcomment{}
We solve \eqref{eq:constrained} via its Lagrangian relaxation.

\paragraph{Uncertainty.}
Let $p_y(x;\mathcal{P}^{(r-1)})$ denote the empirical predictive distribution
over labels obtained from stochastic decoding or prompt perturbations
(as defined in \cref{eq:eq-1}). 
Disagreement-based uncertainty is defined as
\begin{equation}
u(x;\mathcal{P}^{(r-1)})
=
1 -
\max_{y \in \mathcal{Y}}
p_y(x;\mathcal{P}^{(r-1)}),
\label{eq:disagreement-tight}
\end{equation}
which lies in $[0,1]$ and increases when predictions disagree.

\paragraph{Caption complexity.}
Let $\text{toklen}(c)$ denote the token length of caption $c$.
To ensure scale compatibility, we normalize token length by
a fixed upper bound $L_{\max}$:
\[
\ell(c) := \frac{\text{toklen}(c)}{L_{\max}} \in [0,1].
\]

Let $\psi(c) \in \mathbb{R}^d$ denote a fixed embedding of caption $c$,
and let $\text{sim}(\cdot,\cdot)$ denote cosine similarity,
which lies in $[-1,1]$.
To measure redundancy with the existing prompt set,
we use average similarity:
{\footnotesize
\begin{equation}
\begin{aligned}
 \rho(c;\mathcal{P}^{(r-1)})
=
&\frac{1}{|\mathcal{P}^{(r-1)}|}\\
&\sum_{(x',c') \in \mathcal{P}^{(r-1)}}
\frac{1 + \text{sim}(\psi(c),\psi(c'))}{2}
\in [0,1] 
\end{aligned}
\label{eq:similarity}
\end{equation}
}

The anticipated complexity of adding $\hat{c}(x;\mathcal{P}^{(r-1)})$
is then defined as
\begin{equation}
\begin{aligned}
\hat{\kappa}(x;\mathcal{P}^{(r-1)})
&=
\alpha \, \ell(\hat{c}(x;\mathcal{P}^{(r-1)}))+ \\
&\beta \, \rho(\hat{c}(x;\mathcal{P}^{(r-1)});\mathcal{P}^{(r-1)}),
\end{aligned}
\label{eq:complexity-tight}
\end{equation}
where $\alpha,\beta \ge 0$.

Both terms now lie in $[0,1]$, so $\hat{\kappa}(x) \in [0,\alpha+\beta]$,
ensuring scale compatibility with uncertainty.

\paragraph{Lagrangian acquisition score.}
The constrained problem \eqref{eq:constrained}
is optimized via the Lagrangian objective
\begin{equation}
s(x;\mathcal{P}^{(r-1)})
=
u(x;\mathcal{P}^{(r-1)})
-
\lambda_c
\, \hat{\kappa}(x;\mathcal{P}^{(r-1)}),
\label{eq:selection-tight}
\end{equation}
where $\lambda_c \ge 0$ is the dual parameter controlling the trade-off.
This corresponds to maximizing uncertainty subject to a soft
complexity budget.

\paragraph{Active set selection.}
The active set of size $m$ is formed by
\begin{equation}
\mathcal{A}^{(r)}
=
\operatorname{Top\mbox{-}m}
\big\{
(x, s(x;\mathcal{P}^{(r-1)}))
:
x \in \mathcal{D}_u
\big\}.
\label{eq:active-tight}
\end{equation}

This acquisition mechanism therefore selects samples that
(i) are uncertain under the current prompt set,
(ii) introduce concise explanations, and
(iii) minimize semantic redundancy with previously selected captions,
under an explicit Lagrangian relaxation of a constrained uncertainty maximization problem.
\subsection{Density-Tree Boundary Acquisition}
\label{ssec:apt-dtb}

Prior work has shown that clustering in feature embedding space of image encoders can help identify informative samples for AL \cite{coreset, cec}. However, many recent clustering-based AL methods implicitly assume large unlabeled pools and emphasize global coverage or centroid representativeness, which can dilute label budget when expert annotation is expensive, typical in biomedical imaging. Here, we adopt a density-tree (steepest-uphill) construction \cite{silverman1998density} that targets boundary (valley) regions between density clusters in VLM image encoder embedding space, prioritizing decision-ambiguous cases and thereby concentrating constrained expert effort where it most reduces supervision.
% \JMBcomment{I think expert effort is constrained, rather than being limited. Limited may be misconstrued.}

Let $\phi : \mathcal{X} \rightarrow \mathbb{R}^d$ denote a fixed image embedding function, and let $z_i = \phi(x_i)$ for $x_i \in \mathcal{D}_u$. We first construct a $k$-nearest neighbor (kNN) graph using a distance metric $d(\cdot,\cdot)$. Let $N_k(i)$ denote the $k$ nearest neighbors of point $i$.

\paragraph{Local density.}
A rank-based density proxy is defined as
\begin{equation}
\rho_i 
= \left(
\frac{1}{k_\rho} 
\sum_{j \in N_{k_\rho}(i)} 
d(z_i, z_j)
\right)^{-1},
\label{eq:density}
\end{equation}
where higher $\rho_i$ indicates locally denser regions and $k_\rho$ ($k_\rho \le k$) denotes the number of nearest neighbors used to compute the local density proxy. Accordingly, $N_{k_\rho}(i)$ denotes the set of the $k_\rho$ nearest neighbors of $i$ (a subset of $N_k(i)$).

\paragraph{Steepest-uphill tree.}
For each point $i$, define a locality threshold
\begin{equation}
t_i = d\!\left(z_i, z_{i,(k_t)}\right),
\label{eq:threshold}
\end{equation}
where $z_{i,(k_t)}$ is the $k_t$-th nearest neighbor of $i$. 
Candidate parents are
\begin{equation}
\mathcal{C}(i) 
= \{ j \in N_k(i) : d(z_i,z_j) \le t_i \ \wedge \ \rho_j > \rho_i \}.
\label{eq:candidates}
\end{equation}
If $\mathcal{C}(i) \neq \emptyset$, we assign a parent pointer via
\begin{equation}
\pi(i) 
= \arg\max_{j \in \mathcal{C}(i)} 
\frac{\rho_j - \rho_i}{d(z_i,z_j)}.
\label{eq:parent}
\end{equation}
Otherwise, $i$ is a root (local density mode). 
Tracing parent pointers partitions the pool into clusters defined by shared roots.

\paragraph{Boundary score.}
Let $r(i)$ denote the root of $i$. Let $k_b$ denote the neighborhood size used to compute local cluster mixing, with $N_{k_b}(i)$ the set of the $k_b$ nearest neighbors of $i$ ( $k_b \le k$). 

To prioritize samples near cluster-separating regions, we define a boundary score
\begin{equation}
b(i) 
= 1 - 
\frac{1}{k_b}
\sum_{j \in N_{k_b}(i)} 
\mathbf{1}[r(j) = r(i)],
\label{eq:boundary}
\end{equation}
where the indicator equals $1$ when neighbor $j$ belongs to the same density cluster as $i$, and $0$ otherwise. 
Points deep inside clusters have $b(i) \approx 0$, whereas points near cluster interfaces exhibit larger values.

Over-selection of \emph{collective outliers} which are small, isolated groups of points can form spurious low-density clusters due to noise or embedding artifacts \cite{collectiveOutliers}. Such tiny clusters can yield artificially high boundary scores because many neighbors lie outside the clusters, causing the method to waste budget on unstable ``islands'' rather than informative cluster boundaries. We therefore restrict the contribution of very small clusters and limit repeated selections from the same cluster. Details and values of all hyperparameters are provided in \cref{ssec:implementation}.

% To avoid over-selecting \emph{collective outliers}---small, isolated groups that can form spurious clusters due to noise or embedding artifacts \cite{collectiveOutliers}---we apply simple cluster-size safeguards. Such tiny clusters may exhibit inflated boundary scores, wasting query budget on unstable ``islands'' rather than informative boundaries; we therefore cap selections from very small clusters. Hyperparameter values are reported in \cref{ssec:implementation}.

\paragraph{Active set selection.}
 The active set of size $m$ is formed by
\begin{equation}
\mathcal{A}^{(r)} 
= \text{Top-}m \{ (x_i, b(i)) : x_i \in \mathcal{D}_u \},
\label{eq:dts-active}
\end{equation}
thereby selecting samples located near density transitions in embedding space. 

This acquisition strategy targets valley and saddle regions between high-density clusters, where decision ambiguity is likely to concentrate.

\subsection{Annotation}
\label{ssec:annotation}
After selecting the active set $\mathcal{A}^{(r)}$ via the acquisition function, we annotate it using an human-in-the-loop framework designed to minimize manual captioning effort. Rather than writing captions from scratch, we prompt the VLM with the system prompt $S$ and the current prompt set $\mathcal{P}^{(r-1)}$, and apply it to each image $x \in \mathcal{A}^{(r)}$ to generate a candidate caption. Each generated caption is then reviewed and corrected by a domain expert to ensure label correctness and accurate reasoning. We denote the resulting expert-verified set as $\widehat{\mathcal{A}}^{(r)}=\{(x,\tilde{c}) : x\in \mathcal{A}^{(r)}\}$, where $\tilde{c}$ is the corrected caption for image $x$. The prompt set is then updated by appending these new exemplars and removing the corresponding images from the unlabeled pool:
\begin{equation}
\mathcal{P}^{(r)} = \mathcal{P}^{(r-1)} \cup \widehat{\mathcal{A}}^{(r)}, 
\qquad
\mathcal{D}_u \leftarrow \mathcal{D}_u \setminus \mathcal{A}^{(r)}.
\label{eq:prompt_update}
\end{equation}
Following this update, we evaluate the stopping criteria to decide whether to terminate or proceed to the next round.

\subsection{Stopping Criteria}

After each round, once the newly annotated exemplars have been incorporated into the prompt set, we determine whether to continue the active prompting loop using three practical stopping conditions. Specifically, we stop if the mean class accuracy on a held-out validation set reaches a predefined target, if the cumulative number of expert-annotated images exceeds a fixed annotation budget, or if a maximum number of rounds is reached. The procedure terminates as soon as any of these criteria are met.

Details and values of these stopping criteria are provided in \cref{ssec:implementation}.

\subsection{Inference}

Once a stopping criterion is met, we retain the resulting prompt set as the final (optimized) set of annotated exemplars. During inference, each image is prompted to the VLM with the system prompt together with this final prompt set as few-shot context, and the model output is used as the classification prediction for that image.

\section{Experiment Settings}
%========================================================
% CVPR-friendly (single-column) tables: compact + fixed width
% Requires: \usepackage{booktabs,tabularx}
%========================================================
\subsection{Dataset}
\label{ssec:dataset}
We used the same dataset our baseline was evaluated on in \cite{gpt4oISBI2025}. The \emph{Lurcher data} consists of 2-D microscopy images of histologically stained 3-D structures in tissue sections through the cerebellum of 21 mice brains (11 Lurcher mutation, 10 wild-type control group). The classification task involves distinguishing Lurcher mutant mice from the wild type. All the images are captured at low magnification (10x) and stained with Cresyl violet. More details about this dataset can be found in \cite{gpt4oISBI2025}. A four-fold leave-one-out evaluation protocol was used to evalute the proposed methods. In each fold, images from two animals (one per class) were held out as the test set, and images from an additional two animals (one per class) were reserved for validation. From the remaining seventeen animals, an initial seed set was created by labeling ten images from two randomly selected animals (one per class) to initialize the first round. All remaining images from the training animals were treated as the unlabeled pool.

\subsection{Evaluation Metrics}
\label{ssec:eval-metrics}
We report performance at both the image level and the animal level, and we additionally quantify annotation efficiency by the number of expert-annotated images required to reach the stopping criteria.

\textbf{Mean class accuracy (\%)} The mean class accuracy is computed by first measuring the percentage of images from that class that are correctly classified for each class, and then averaging these per-class accuracies across the two classes.

\textbf{Animal accuracy (\%)} Animal accuracy is also reported as it reflects the relevant decision making process. For each animal, we aggregate the predicted labels across its images using majority voting. The animal is then assigned the class receiving more than $50\%$ of the image-level predictions. Animal-level accuracy is then computed as the percentage of test animals whose aggregated label matches the ground truth.

\textbf{Number of manually annotated images} Number of images that are manually corrected and verified up to the method termination is reported to measure annotation efficiency. Fewer annotated images indicate lower expert time and effort and higher annotation efficiency.

\subsection{Baseline}

% We use the APT-USF framework~\cite{gpt4oISBI2025} as our baseline. APT-USF follows the same annotation process as our proposed variants but differs in  its acquisition strategy: at each round, it selects the active set by uniform sampling from the unlabeled pool rather than by a score-based acquisition function.

% We use the APT-USF framework as our baseline. In APT-USF, the few-shot context is provided by a fixed size prompt set of 36 images, and at each round it selects the active set by uniform sampling from the unlabeled pool.  More details on the experiment settings of APT-USF can be found in \cite{gpt4oISBI2025}.

We use the APT-USF framework as our baseline. In APT-USF, the active set is selected each round by uniform sampling from the unlabeled pool, and the prompt set is expanded iteratively until it reaches a fixed size of 36 images. More details on the experiment settings of APT-USF can be found in \cite{gpt4oISBI2025}.

Across all methods, we use GPT-4o via the OpenAI API, motivated by (i) strong reported performance on challenging histopathology image classification tasks in prior studies ~\cite{LMMsForPathSPIE, gpt4oISBI2025, gpt4vForNeuroScience}, and (ii) the need for a consistent model across the baseline and proposed methods for a fair comparison.

\subsection{Implementation Details}
\label{ssec:implementation}

\paragraph{Hyperparameters}
All method use the same number of exemplars per round: each round adds $m{=}10$ annotated exemplars to the prompt set. To simulate performance evaluation under low-resource constraints, we restrict acquisition to a fixed candidate subset $C^{(r)} \subset D_u$ with $|C^{(r)}|{=}100$ at each round for all proposed methods. We use the validation-accuracy stopping criterion described in \cref{ssec:eval-metrics} with a target threshold of $80\%$, chosen in consultation with a neuroscience domain-expert (who is also a co-author) and supported by pilot experiments. 
% As the prompt set grows across rounds, the input context length increases, and prior studies have shown that model performance can degrade when reasoning must rely on information embedded in long contexts \cite{lost_in_the_middle}. 
To keep the prompt context length within a range where model performance remains stable \cite{lost_in_the_middle}, we additionally impose a round limit of $R_{\max}{=}7$.

\paragraph{Uncertainty estimation}
For acquisition functions that require uncertainty (\cref{ssec:apt-u} and \cref{ssec:apt-ca}), we estimate uncertainty via stochastic decoding: for each candidate image we generate $T{=}5$ stochastic outputs using temperature-based sampling \cite{self-consistency}, construct an empirical label distribution from these samples (\cref{eq:eq-1}), and compute the corresponding uncertainty scores (e.g., entropy or disagreement).

\paragraph{Complexity-aware uncertainty acquisition}
The complexity-aware uncertainty acquisition requires (i) a tokenizer to compute caption length and (ii) a text embedding model to measure semantic redundancy between a candidate caption and the existing prompt-set captions (\cref{ssec:apt-ca}). Concretely, we compute $\text{toklen}(\cdot)$ in \cref{eq:complexity-tight} using the \texttt{cl100k\_base} encoding \cite{tiktoken}, which matches the tokenization used by GPT-4 models.

To compute the similarity term in \cref{eq:complexity-tight}, we embed captions with a lightweight Sentence-Transformer encoder \cite{minilmv2}, i.e., $\psi(c)\in\mathbb{R}^d$. We use cosine similarity $\text{sim}(\psi(c_1),\psi(c_2))$ to approximate semantic overlap between captions. We choose this text encoder for simplicity and computational efficiency, and we found it sufficient for this purpose in our experiments.

The weighting coefficients $(\alpha,\beta)$ control the relative penalty on caption length and redundancy in \cref{eq:complexity-tight}, while $\lambda_c$ controls the trade-off between uncertainty and the complexity penalty in \cref{eq:selection-tight}. We selected $(\alpha=0.01,\beta=0.1,\lambda_c=0.5)$ via a small validation sweep and fixed them across all experiments.

\paragraph{Density-tree boundary acquisition}
% The Density-Tree Boundary (DTB) acquisition in \cref{ssec:apt-dtb} depends on the embedding function $\phi(\cdot)$ used to compute $z=\phi(x)$ for constructing the $k$NN graph and density-tree structure. We use BioMedCLIP \cite{biomedclip} as the embedding backbone for the main results and report comparisons with alternative embedding models in \cref{sec:ablation}. We use cosine distance throughout.

The Density-Tree Boundary (DTB) acquisition in \cref{ssec:apt-dtb} depends on the embedding function $\phi(\cdot)$ used to compute $z=\phi(x)$ and the distance metric $d(\cdot,\cdot)$ used to define local neighborhoods for density estimation and density-tree construction. In all DTB computations we use \emph{cosine distance}, i.e., one minus cosine similarity, to measure distances between embedding vectors. For the main results, BioMedCLIP \cite{biomedclip} image encoder is the embedding backbone and report comparisons with alternative embedding models in \cref{ssec:ablation}.

We set neighborhood and cluster-related hyperparameters to preserve (i) locality for estimating $\rho_i$ (Eq.~\eqref{eq:density}) and boundary mixing $b(i)$ (Eq.~\eqref{eq:boundary}), (ii) robustness to spurious edges and collective outliers \cite{collectiveOutliers}, and (iii) diversity across density clusters within each round. These hyperparameters were selected via a small tuning study using unsupervised diagnostics on candidate-pool geometry and held fixed across all experiments.

We build the DTB neighborhood graph using $k=20$ neighbors. Local density is estimated with $k_\rho=10$ neighbors (Eq.~\eqref{eq:density}), and the adaptive locality threshold uses $k_t=10$ (Eq.~\eqref{eq:threshold}). Boundary mixing is computed over $k_b=8$ neighbors (Eq.~\eqref{eq:boundary}), providing a stable local estimate. To avoid over-selecting unstable regions and collective outliers, we treat clusters with fewer than $m_{\min}=6$ points as ``tiny'' and cap selections to at most one sample per tiny cluster and at most one sample per cluster overall. 
% Finally, tiny clusters are only eligible for selection when they appear boundary-like by requiring $b(i)\ge b_{\min}=0.6$, which reduces spending query budget on collective outlier ``islands''.

% all other density-tree hyper-parameters follow the definitions in Sec.~\ref{sec:density_tree_acq} and are held constant across folds.
\section{Results}
\label{sec:results}
We compare the baseline APT-USF (random acquisition) with three proposed variants that differ in the acquisition function: APT-U (uncertainty-guided), APT-CA (complexity-aware uncertainty), and APT-DTB (density-tree boundary sampling) across four folds of the Lurcher dataset described in \ref{ssec:dataset}.
% Results are summarized in Tables~\ref{tab:lurcher_efficiency} and~\ref{tab:lurcher_accuracy}.

Table~\ref{tab:lurcher_efficiency} reports annotation efficiency, measured as the number of expert-annotated images required to first reach the validation target of $80\%$ mean class accuracy, along with the resulting animal-level test accuracy (majority vote). All three proposed acquisition functions achieve $100\%$ animal-level test accuracy, while requiring only a small number of expert annotations (at most 35 images per fold). Notably, this performance is obtained under a strict low-resource constraint in which acquisition is performed from a candidate subset of only $100$ unlabeled images per round. Among the compared methods, APT-U is the most annotation-efficient, reaching the stopping criterion with the fewest annotated images while maintaining perfect animal-level accuracy.

Table~\ref{tab:lurcher_accuracy} reports image-level test performance in terms of mean class accuracy. All proposed variants improve over the random-acquisition baseline, indicating that the acquisition function substantially accelerates learning under limited annotation budgets. In particular, APT-CA achieves the best average test performance, improving mean class accuracy by approximately $8$ points over APT (88.6 vs.\ 80.1), with APT-U and APT-DTB also yielding consistently strong gains. Overall, these results highlight the effectiveness of uncertainty- and structure-aware acquisition for annotation-efficient tuning in low-data microscopy image classification.
%---------------------------%
% Table 1: Efficiency to Target + Animal-level test
%---------------------------%
\begin{table}[t]
\caption{\textbf{Annotation efficiency.}
Each method terminates when it first reaches the validation target (80\% mean class accuracy). We report the number of expert-annotated images (mean$\pm$std across four folds) required to stop and the resulting animal-level test accuracy (majority vote). Bold indicates the best (lowest) annotation cost.}

\label{tab:lurcher_efficiency}
\centering
\setlength{\tabcolsep}{4pt}
\renewcommand{\arraystretch}{1.08}
\begin{tabularx}{\columnwidth}{@{}lcc@{}}
\toprule
Method & \# Images Annotated\ $\downarrow$ & Animal Accuracy \\
\midrule
APT & $36\pm0$ & 87.5\% \\
APT-U    & $\mathbf{20\pm0}$ & 100\% \\
APT-CA  & $23\pm5$ & 100\% \\
APT-DTB  & $25\pm10$ & 100\% \\
\bottomrule
\end{tabularx}
\end{table}

%---------------------------%
% Table 2: Image-level generalization summary
%---------------------------%
\begin{table}[t]
\caption{\textbf{Test performance.}
After each method meets the validation criterion and stops, we evaluate on the held-out test split. We report class-wise image-level accuracy (mean$\pm$std across four folds) and animal-level accuracy via majority vote. Bold indicates the best test performance.}
\label{tab:lurcher_accuracy}
\centering
\setlength{\tabcolsep}{3.5pt}
\renewcommand{\arraystretch}{1.0}
\begin{tabularx}{\columnwidth}{@{}lcc@{}}
\toprule
Method & Mean Class Accuracy & Animal Accuracy \\
\midrule
APT & $80.1\pm14.7$\% & 87.5\% \\
APT-U    & $87.8\pm6.0$\%  & 100\% \\
APT-CA  & $\mathbf{88.6\pm4.4}$\% & 100\% \\
APT-DTB  & $86.5\pm4.3$\%  & 100\% \\
\bottomrule
\end{tabularx}
\end{table}

\subsection{Ablation Study: Effect of the Embedding Backbone}
\label{ssec:ablation}

\begin{table}[t]
\caption{\textbf{Effect of the embedding backbone on APT-DTB (Lurcher, 3 folds).}
APT-DTB constructs a density-tree in the embedding space induced by the image encoder. We report the mean$\pm$std number of expert-annotated images required to first reach the validation target (80\% mean class accuracy), the resulting animal-level test accuracy (majority vote).}

\label{tab:ablation_embeddings}
\centering
\setlength{\tabcolsep}{4pt}
\renewcommand{\arraystretch}{1.08}
\begin{tabularx}{\columnwidth}{@{}lcc@{}}
\toprule
Embedding\\ backbone &
\# images annotated $\downarrow$ &
Animal accuracy \\
\midrule
CLIP        & $40 \pm 28$ & 100\% \\
BioMedCLIP  & $\mathbf{27 \pm 9}$  & 100\% \\
MedSigLIP  &  $40 \pm 28$  & 100\% \\
\bottomrule
\end{tabularx}
\end{table}

% We next analyze how individual design choices affect performance under the same evaluation protocol. In particular, we study factors that directly influence sample selection quality in our acquisition functions.

% \subsection{Effect of the Embedding Backbone}
% \label{ssec:ablation-embedding}
The density-tree boundary acquisition (APT-DTB) operates in an embedding space, where images are mapped to feature vectors and clustered via a density-based (steepest-uphill) construction (\cref{ssec:apt-dtb}). As a result, the embedding backbone $\phi(\cdot)$ is a key component: it defines the geometry of the latent space in which density clusters and boundary regions are identified, and therefore directly impacts which samples are selected for annotation.

To assess this sensitivity, we instantiate APT-DTB with three embedding backbones spanning general-purpose to medical-domain pretraining.

\textbf{CLIP} is a contrastively trained vision--language model learned from \(\sim\)400M (400 million) web-scale image--text pairs. We use the image encoder from ViT-B/16 variant (\(\sim\)86M parameters) as a strong domain-agnostic baseline for defining the embedding space \cite{clip}.

\textbf{BioMedCLIP} follows the same CLIP-style objective but is pretrained on PMC-15M (15M biomedical figure--caption pairs), using a ViT-B/16 image encoder together with a PubMedBERT text encoder (\(\sim\)110M parameters), thereby biasing representations toward biomedical semantics \cite{biomedclip}.

\textbf{MedSigLIP} is a medical-domain SigLIP variant with a 400M-parameter vision encoder and 400M-parameter text encoder, trained on diverse de-identified 33M medical image--text pairs spanning modalities including histopathology \cite{medsiglip}. 

As shown in Table~\ref{tab:ablation_embeddings}, BioMedCLIP yields the highest annotation efficiency for APT-DTB, reaching the validation target with fewer expert-annotated images than both CLIP and MedSigLIP, while all three backbones enable $100\%$ test animal accuracy. 
This outcome is contrary to our initial expectation that MedSigLIP would dominate due to its larger encoder size and medical training dataset. 
% A plausible explanation is that BioMedCLIP's contrastive alignment on biomedical figure--caption pairs induces a more discriminative and locally consistent embedding features for our microscopy images, which benefits density-based boundary selection.

% \input{sec/6_Ablations}
\section{Conclusion}

We proposed a target-driven active prompt tuning framework for microscopy image classification that constructs a compact \emph{prompt set} while minimizing expert annotation effort. Instead of treating the prompt set size as a fixed design choice, our approach iteratively expands the prompt set and terminates once a validation performance target is achieved. Within this framework, we studied three complementary acquisition functions that differ in how the next set of unlabeled images are chosen for expert verification. Empirically, these acquisition strategies improved annotation efficiency over random selection while maintaining 100\% test accuracy, demonstrating that our methods can substantially reduce expert effort in low-resource microscopy classification.

There are a few directions for future work. First, evaluating the proposed methods on additional diverse datasets will be important to assess generalizability. Second, evaluating the proposed framework across multiple general-purpose and medical pre-trained VLMs would clarify how acquisition behaviors vary with model capacity, pretraining, and would improve the practical applicability of the framework.

\section{Acknowledgments}
\label{sec:acknowledgments}

This work was supported by National Science Foundation grants (\#1513126, \#1746511, \#1926990) and a Florida High Tech Corridor grant (ENG203, \#20-10) to SRC Biosciences and the University of South Florida. We also thank Dr.\ Yaroslav Kolinko (Faculty of Medicine in Pilsen, Charles University, Czech Republic) for providing the dataset used in this work.
{
    \small
    \bibliographystyle{ieeenat_fullname}
    \bibliography{main}

@inproceedings{wsi-mil,
  title={Dual-stream multiple instance learning network for whole slide image classification with self-supervised contrastive learning},
  author={Li, Bin and Li, Yin and Eliceiri, Kevin W},
  booktitle={Proceedings of the IEEE/CVF conference on computer vision and pattern recognition},
  pages={14318--14328},
  year={2021}
}

@book{stereo-book,
author = {Mouton, Peter R.},
year = {2011},
month = {08},
pages = {},
title = {Unbiased Stereology: A Concise Guide},
isbn = {978-0801899850},
publisher = {Johns Hopkins University Press}
}

@inproceedings{morera2023mimo,
  author={Morera, H. and Dave, P. and Alahmari, S. and Kolinko, Y. and Hall, L.O. and Goldgof, D. and Mouton, P.R.},
  booktitle={2023 IEEE 36th International Symposium on Computer-Based Medical Systems (CBMS)}, 
  title={MIMO YOLO - A Multiple Input Multiple Output Model for Automatic Cell Counting}, 
  year={2023},
  volume={},
  number={},
  pages={827-831},
  doi={10.1109/CBMS58004.2023.00327}}

@article{morera20x,
  title={A novel deep learning-based method for automatic stereology of microglia cells from low magnification images},
  author={Morera, Hunter and Dave, Palak and Kolinko, Yaroslav and Alahmari, Saeed and Anderson, Aidan and Denham, Grant and Davis, Chloe and Riano, Juan and Goldgof, Dmitry and Hall, Lawrence O. and others},
  journal={Neurotoxicology and Teratology},
  volume={102},
  pages={107336},
  year={2024},
  publisher={Elsevier}}

@inproceedings{palakmimounet,
       author = {{Dave}, Palak and {Kolinko}, Yaroslav and {Morera}, Hunter and {Allen}, Kurtis and {Alahmari}, Saeed and {Goldgof}, Dmitry and {Hall}, Lawrence O. and {Mouton}, Peter R.},
        title = "{MIMO U-Net: efficient cell segmentation and counting in microscopy image sequences}",
    booktitle = {Society of Photo-Optical Instrumentation Engineers (SPIE) Conference Series},
         year = {2023},
       editor = {{Tomaszewski}, J.E. and {Ward}, A.D.}
}

@article{why-ICL,
  title={Why can gpt learn in-context? language models implicitly perform gradient descent as meta-optimizers},
  author={Dai, Damai and Sun, Yutao and Dong, Li and Hao, Yaru and Ma, Shuming and Sui, Zhifang and Wei, Furu},
  journal={arXiv preprint arXiv:2212.10559},
  year={2022}
}

@article{ICL-examples,
  title={What makes good examples for visual in-context learning?},
  author={Zhang, Yuanhan and Zhou, Kaiyang and Liu, Ziwei},
  journal={Advances in Neural Information Processing Systems},
  volume={36},
  year={2024}
}

@inproceedings{margin_AL,
author = {Balcan, Maria-Florina and Broder, Andrei and Zhang, Tong},
title = {Margin based active learning},
year = {2007},
isbn = {9783540729259},
publisher = {Springer-Verlag},
address = {Berlin, Heidelberg},
booktitle = {Proceedings of the 20th Annual Conference on Learning Theory},
pages = {35–50},
numpages = {16},
location = {San Diego, CA, USA},
series = {COLT'07}
}

@article{coop,
  title={Learning to prompt for vision-language models},
  author={Zhou, Kaiyang and Yang, Jingkang and Loy, Chen Change and Liu, Ziwei},
  journal={International journal of computer vision},
  volume={130},
  number={9},
  pages={2337--2348},
  year={2022},
  publisher={Springer}
}

@inproceedings{cocoop,
  title={Conditional prompt learning for vision-language models},
  author={Zhou, Kaiyang and Yang, Jingkang and Loy, Chen Change and Liu, Ziwei},
  booktitle={Proceedings of the IEEE/CVF conference on computer vision and pattern recognition},
  pages={16816--16825},
  year={2022}
}

@inproceedings{residualprompttuning,
  title={Residual prompt tuning: Improving prompt tuning with residual reparameterization},
  author={Razdaibiedina, Anastasiia and Mao, Yuning and Khabsa, Madian and Lewis, Mike and Hou, Rui and Ba, Jimmy and Almahairi, Amjad},
  booktitle={Findings of the Association for Computational Linguistics: ACL 2023},
  pages={6740--6757},
  year={2023}
}

@inproceedings{bang2024active,
  title={Active prompt learning in vision language models},
  author={Bang, Jihwan and Ahn, Sumyeong and Lee, Jae-Gil},
  booktitle={Proceedings of the IEEE/CVF Conference on Computer Vision and Pattern Recognition},
  pages={27004--27014},
  year={2024}
}

@inproceedings{LMMsForPathSPIE,
author = {Caleb Heinzman and Huazhang Guo and Mai He and Ye Duan},
title = {{LMMs for histopathology: zero- and few-shot patch classification with GPT and Gemini models}},
booktitle = {Ninth International Conference on Advances in Image Processing (ICAIP 2025)},
pages = {140170T},
year = {2026},
doi = {10.1117/12.3098947},
URL = {https://doi.org/10.1117/12.3098947}
}

@inproceedings{active-prompt_acl_2024,
    title = "Active Prompting with Chain-of-Thought for Large Language Models",
    author = "Diao, Shizhe  and
      Wang, Pengcheng  and
      Lin, Yong  and
      Pan, Rui  and
      Liu, Xiang  and
      Zhang, Tong",
    booktitle = "Proceedings of the 62nd Annual Meeting of the Association for Computational Linguistics (Volume 1: Long Papers)",
    month = aug,
    year = "2024",
    publisher = "Association for Computational Linguistics",
    url = "https://aclanthology.org/2024.acl-long.73/",
    doi = "10.18653/v1/2024.acl-long.73",
    pages = "1330--1350",
}

@inproceedings{COT,
 author = {Wei, Jason and Wang, Xuezhi and Schuurmans, Dale and Bosma, Maarten and Ichter, Brian and Xia, Fei and Chi, Ed and Le, Quoc V and Zhou, Denny},
 booktitle = {Advances in Neural Information Processing Systems},
 editor = {S. Koyejo and S. Mohamed and A. Agarwal and D. Belgrave and K. Cho and A. Oh},
 pages = {24824--24837},
 publisher = {Curran Associates, Inc.},
 title = {Chain-of-Thought Prompting Elicits Reasoning in Large Language Models},
 url = {https://proceedings.neurips.cc/paper_files/paper/2022/file/9d5609613524ecf4f15af0f7b31abca4-Paper-Conference.pdf},
 volume = {35},
 year = {2022}
}

@inproceedings{entropy,
  title={Entropy-based active learning for object recognition},
  author={Holub, Alex and Perona, Pietro and Burl, Michael C},
  booktitle={ IEEE computer society conference on computer vision and pattern recognition workshops},
  pages={1--8},
  year={2008},
  organization={IEEE}
}

@article{self-consistency,
  title={Self-consistency improves chain of thought reasoning in language models},
  author={Wang, Xuezhi and Wei, Jason and Schuurmans, Dale and Le, Quoc and Chi, Ed and Narang, Sharan and Chowdhery, Aakanksha and Zhou, Denny},
  journal={arXiv preprint arXiv:2203.11171},
  year={2022}
}

@inproceedings{
coreset,
title={Active Learning for Convolutional Neural Networks: A Core-Set Approach},
author={Ozan Sener and Silvio Savarese},
booktitle={International Conference on Learning Representations},
year={2018},
url={https://openreview.net/forum?id=H1aIuk-RW},
}

@inproceedings{cec,
  title={Active learning for vision-language models},
  author={Safaei, Bardia and Patel, Vishal M},
  booktitle={2025 IEEE/CVF Winter Conference on Applications of Computer Vision (WACV)},
  pages={4902--4912},
  year={2025},
  organization={IEEE}
}

@ARTICLE{gpt4vForNeuroScience,
  title    = "Evaluating the efficacy of few-shot learning for {GPT-4Vision} in
              neurodegenerative disease histopathology: A comparative analysis
              with convolutional neural network model",
  author   = "Ono, Daisuke and Dickson, Dennis W and Koga, Shunsuke",
  journal  = "Neuropathol Appl Neurobiol",
  volume   =  50,
  number   =  4,
  pages    = "e12997",
  month    =  aug,
  year     =  2024,
  address  = "England"
}

@INPROCEEDINGS{gpt4oISBI2025,
  author={Abhiram Kandiyana and Peter R. Mouton and Yaroslav Kolinko and Lawrence O. Hall and Dmitry Goldgof},
  booktitle={2025 IEEE 22nd International Symposium on Biomedical Imaging (ISBI)}, 
  title={Active Prompt Tuning Enables GPT-4O to do Efficient Classification of Microscopy Images}, 
  year={2025},
  volume={},
  number={},
  pages={01-05},
  doi={10.1109/ISBI60581.2025.10981114}}

@inproceedings{clip,
  title={Learning transferable visual models from natural language supervision},
  author={Radford, Alec and Kim, Jong Wook and Hallacy, Chris and Ramesh, Aditya and Goh, Gabriel and Agarwal, Sandhini and Sastry, Girish and Askell, Amanda and Mishkin, Pamela and Clark, Jack and others},
  booktitle={International conference on machine learning},
  pages={8748--8763},
  year={2021},
  organization={PMLR}
}

@article{biomedclip,
  title={BiomedCLIP: a multimodal biomedical foundation model pretrained from fifteen million scientific image-text pairs},
  author={Zhang, Sheng and Xu, Yanbo and Usuyama, Naoto and Xu, Hanwen and Bagga, Jaspreet and Tinn, Robert and Preston, Sam and Rao, Rajesh and Wei, Mu and Valluri, Naveen and others},
  journal={arXiv preprint arXiv:2303.00915},
  year={2023}
}

@article{medsiglip,
  title={MedGemma Technical Report},
  author={Sellergren, Andrew and Kazemzadeh, Sahar and Jaroensri, Tiam and Kiraly, Atilla and Traverse, Madeleine and Kohlberger, Timo and Xu, Shawn and Jamil, Fayaz and Hughes, Cían and Lau, Charles and others},
  journal={arXiv preprint arXiv:2507.05201},
  year={2025}
}

@techreport{settles2009,
  author      = {Burr Settles},
  title       = {Active Learning Literature Survey},
  institution = {University of Wisconsin--Madison},
  number      = {1648},
  year        = {2009}
}

@inproceedings{lewis1994,
  author    = {David D. Lewis and William A. Gale},
  title     = {A Sequential Algorithm for Training Text Classifiers},
  booktitle = {Proceedings of SIGIR},
  year      = {1994}
}

@article{tong2001,
  author  = {Simon Tong and Daphne Koller},
  title   = {Support Vector Machine Active Learning with Applications to Text Classification},
  journal = {Journal of Machine Learning Research},
  volume  = {2},
  pages   = {45--66},
  year    = {2001}
}

@inproceedings{BADGE,
title={Deep Batch Active Learning by Diverse, Uncertain Gradient Lower Bounds},
author={Jordan T. Ash and Chicheng Zhang and Akshay Krishnamurthy and John Langford and Alekh Agarwal},
booktitle={International Conference on Learning Representations},
year={2020},
url={https://openreview.net/forum?id=ryghZJBKPS}
}

@inproceedings{minilmv2,
  title={Minilmv2: Multi-head self-attention relation distillation for compressing pretrained transformers},
  author={Wang, Wenhui and Bao, Hangbo and Huang, Shaohan and Dong, Li and Wei, Furu},
  booktitle={Findings of the Association for Computational Linguistics: ACL-IJCNLP 2021},
  pages={2140--2151},
  year={2021}
}

@misc{tiktoken,
  title        = {tiktoken: Fast BPE tokeniser for OpenAI models},
  howpublished = {\url{https://github.com/openai/tiktoken}},
  note         = {Accessed 2026-02-22}
}

@inproceedings{collectiveOutliers,
  title={Mind your outliers! investigating the negative impact of outliers on active learning for visual question answering},
  author={Karamcheti, Siddharth and Krishna, Ranjay and Fei-Fei, Li and Manning, Christopher D},
  booktitle={Proceedings of the 59th Annual Meeting of the Association for Computational Linguistics and the 11th International Joint Conference on Natural Language Processing (Volume 1: Long Papers)},
  pages={7265--7281},
  year={2021}
}

@article{lost_in_the_middle,
    title = "Lost in the Middle: How Language Models Use Long Contexts",
    author = "Liu, Nelson F.  and
      Lin, Kevin  and
      Hewitt, John  and
      Paranjape, Ashwin  and
      Bevilacqua, Michele  and
      Petroni, Fabio  and
      Liang, Percy",
    journal = "Transactions of the Association for Computational Linguistics",
    volume = "12",
    year = "2024",
    address = "Cambridge, MA",
    publisher = "MIT Press",
    url = "https://aclanthology.org/2024.tacl-1.9/",
    doi = "10.1162/tacl_a_00638",
    pages = "157--173"
}

@book{silverman1998density,
  author    = {Bernard W. Silverman},
  title     = {Density Estimation for Statistics and Data Analysis},
  year      = {1998},
  edition   = {1},
  publisher = {Chapman \& Hall/CRC},
  doi       = {10.1201/9781315140919}
}

@book{mouton2014neurostereology,
  title={Neurostereology: unbiased stereology of neural systems},
  author={Mouton, Peter R},
  year={2014},
  publisher={John Wiley \& Sons}
}
}

\end{document}